\documentclass{edm_article}

\usepackage[utf8]{inputenc} 
\usepackage[T1]{fontenc}    
\usepackage{hyperref}       
\usepackage{url}            
\usepackage{booktabs}       
\usepackage{amsfonts}       
\usepackage{amsmath}        
\usepackage{amsfonts}       
\usepackage{nicefrac}       
\usepackage{microtype}      
\usepackage{xcolor}         

\usepackage{amsfonts}
\usepackage{commath}
\usepackage{cleveref}
\usepackage{placeins}
\usepackage{booktabs}
\usepackage{enumitem}
\usepackage{comment}  
\usepackage{graphicx}       

\begin{document}

\title{Faster, Cheaper, More Accurate: Specialised Knowledge Tracing Models Outperform LLMs}

\numberofauthors{4}
\author{
\alignauthor
Prarthana Bhattacharyya\footnotemark[1]\\
       \affaddr{Eedi}\\
       \email{prarthana.bhattacharyya\\@eedi.com}
\alignauthor
Joshua Mitton\footnotemark[1]\\
       \affaddr{Eedi}\\
       \email{joshua.mitton\\@eedi.com}
\alignauthor
Ralph Abboud\\
       \affaddr{Learning Engineering Virtual Institute}\\
       \email{rabboud@levimath.org}
\and
\alignauthor
Simon Woodhead\\
       \affaddr{Eedi}\\
       \email{simon.woodhead@eedi.com}
}

\maketitle
\footnotetext[1]{Equal contribution}
\begin{abstract}
Predicting future student responses to questions is particularly valuable for educational learning platforms where it enables effective interventions. One of the key approaches to do this has been through the use of knowledge tracing (KT) models. These are small, domain-specific, temporal models trained on student question-response data. KT models are optimised for high accuracy on specific educational domains and have fast inference and scalable deployments. The rise of Large Language Models (LLMs) motivates us to ask the following questions: (1) How well can LLMs perform at predicting students' future responses to questions? (2) Are LLMs scalable for this domain? (3) How do LLMs compare to KT models on this domain-specific task? In this paper, we compare multiple LLMs and KT models across predictive performance, deployment cost, and inference speed to answer the above questions. We show that KT models outperform LLMs with respect to accuracy and F1 scores on this domain-specific task. Further, we demonstrate that LLMs are orders of magnitude slower than KT models and cost orders of magnitude more to deploy. This highlights the importance of domain-specific models for education prediction tasks and the fact that current closed source LLMs should not be used as a universal solution for all tasks.
\end{abstract}


\section{Introduction}
\label{sec:introduction}

Predicting student performance is a fundamental and well studied task in educational data mining. The performance of predictive models has important implications for adaptive learning systems and personalised instruction. In mathematics education, students often exhibit recurring patterns of misunderstanding that lead to predictable errors~\cite{brown1978diagnostic}. If these misconceptions can be anticipated, educators can adapt their instruction and learning platforms can deliver more targeted support. This student misconception identification challenge can be framed as a prediction problem: given a student's response history to past questions, the goal is to forecast their performance on future questions.

Knowledge Tracing (KT) models have been developed specifically to address this challenge~\cite{corbett1994knowledge}. These compact, domain-specific models are trained on student interaction data. They are optimised for high accuracy, fast inference, and scalable deployment. Recent advances include Deep Knowledge Tracing (DKT)~\cite{piech2015deep}, which applies recurrent neural networks to model student knowledge states, and Self-Attentive Knowledge Tracing (SAKT)~\cite{pandey2019self}, which leverages attention mechanisms to capture dependencies in student response sequences.

The recent success of Large Language Models (LLMs) across diverse tasks, including mathematical reasoning and problem solving~\cite{openai2023gpt4}, raises a natural question: can general-purpose LLMs match or exceed the performance of specialised KT models on student response prediction? This question is particularly relevant given that predicting student performance on mathematics problems requires both reasoning ability and mathematical understanding. It remains unclear whether these LLM capabilities transfer effectively to modelling individual student knowledge states over time. It is also unclear whether LLMs can do so at a cost and speed suitable for real-world deployment.

In this paper, we present a systematic comparison of KT models and LLMs to address three research questions: (1) How well can LLMs predict students' future responses to questions? (2) Are LLMs scalable for this domain in terms of latency and cost? (3) How do LLMs compare to specialised KT models on this task? We evaluate models on a binary classification task, where the model is tasked to predict whether a student will answer correctly or incorrectly. We do this comparison across three dimensions: predictive performance, inference latency, and deployment cost.

\begin{figure*}[thb!]
    \centering
    \includegraphics[width=0.7\textwidth]{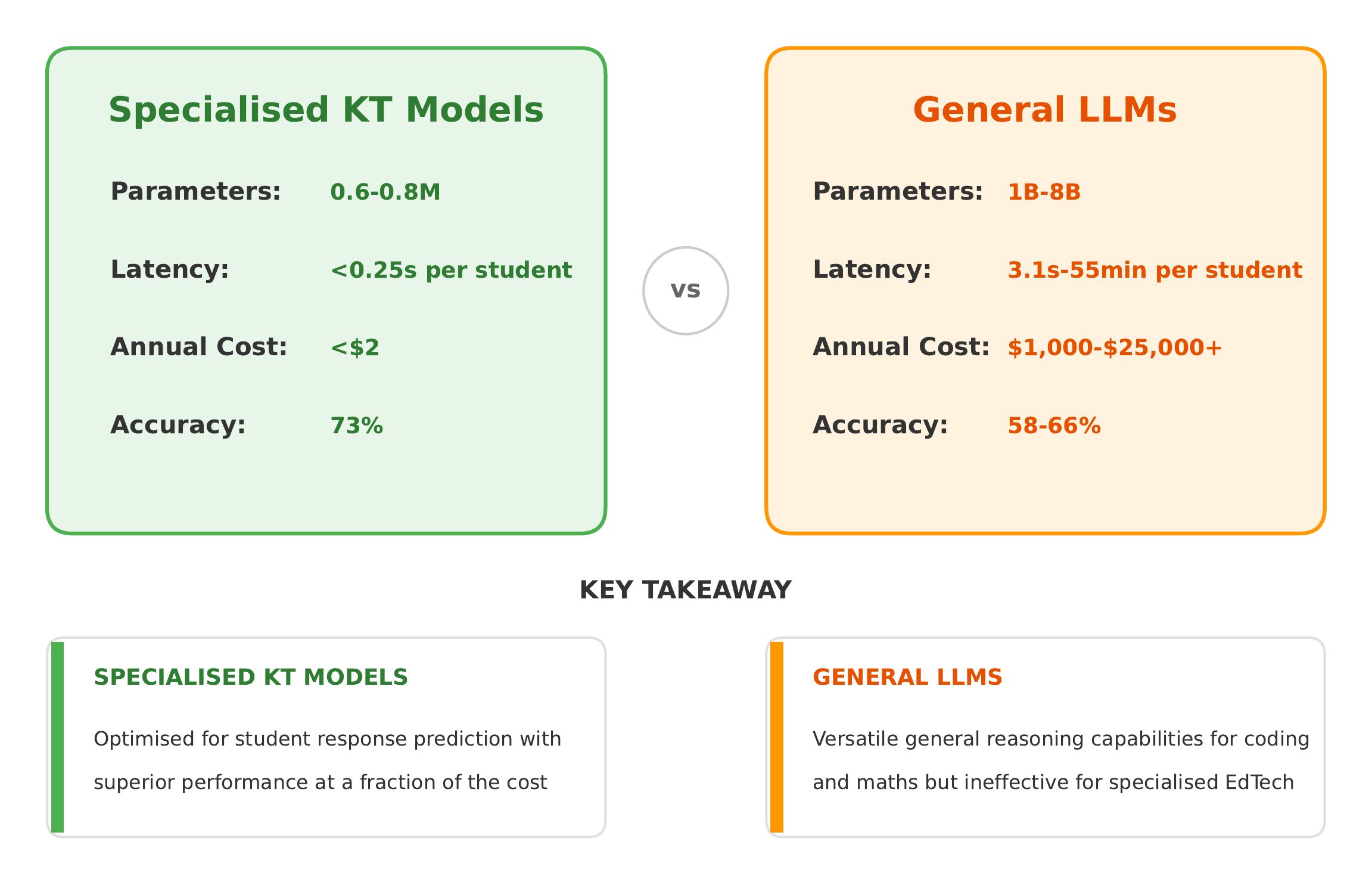}
    \caption{Comparing specialised Knowledge Tracing (KT) models with Large Language Models (LLMs) for students' future performance predictions.}
    \label{fig:intro}
\end{figure*}

Our findings demonstrate that specialised KT models consistently outperform LLMs across all three dimensions, as compared in \Cref{fig:intro}. KT models achieve higher accuracy and F1 scores, are orders of magnitude faster (sub-second vs.\ minutes per student), and cost 600-12,000 times less to deploy at scale. These results highlight the continued importance of domain-specific architectures for educational prediction tasks and caution against adopting general-purpose LLMs as a universal solution in this domain.

\section{Related Work}
\label{sec:related_work}

\subsection{Knowledge Tracing}
Knowledge Tracing (KT) aims to model student knowledge states over time based on their interaction history with learning materials. The foundational approach, Bayesian Knowledge Tracing (BKT), models student knowledge as a latent binary variable that evolves according to learning and forgetting parameters ~\cite{corbett1994knowledge}. Deep Knowledge Tracing (DKT) introduced recurrent neural networks to this domain, enabling automatic feature learning from student response sequences~\cite{piech2015deep}. DKT demonstrated improvements over BKT and sparked deep learning approaches to knowledge tracing. However, DKT has been critiqued for limited interpretability and inconsistent predictions across similar knowledge states~\cite{yeung2018addressing}. More recently, attention-based models have achieved state-of-the-art performance on KT benchmarks. Self-Attentive Knowledge Tracing (SAKT) applies transformer-style self-attention to capture long-range dependencies in student response sequences~\cite{pandey2019self}. Subsequent work has extended this approach with richer question representations~\cite{ghosh2020context} and graph-based skill relationships~\cite{nakagawa2019graph}. In this work, we benchmark against DKT and SAKT as representative examples of recurrent and attention-based KT models, respectively.

\subsection{LLMs for Student Modelling}
Large Language Models have been increasingly applied to educational tasks, including automated essay scoring~\cite{mizumoto2023exploring}, generating pedagogical feedback~\cite{dai2023can}, and mathematical problem solving~\cite{imani2023mathprompter}. Several studies have examined LLMs' ability to model student understanding. LLMs have been shown to infer student misconceptions and adapt teaching strategies better than simple baselines, but worse than methods that explicitly model misconceptions~\cite{ross2024toward}. Furthermore, LLMs struggle to identify incorrect reasoning containing misconceptions more than they do correct reasoning~\cite{sonkar2024malalgoqa}. Recent work has explored combining language models with knowledge tracing. Scarlatos et al.~\cite{scarlatos2025exploring} demonstrated that hybrid approaches can lead to better estimates of student knowledge states than KT-only methods in dialogue settings. Further, LLMs have been used to generate synthetic reasoning traces containing realistic errors~\cite{ross2025learning} and to generate plausible misconceptions given dialogue data~\cite{mitton2026misconceptiondiagnosis}.
These directions suggest potential synergies between LLMs and KT systems, though systematic comparisons of their stand-alone performance on student response prediction remain unexplored.

\subsection{Comparing KT Models with LLMs}
Despite growing interest in LLMs for education, their application to student performance prediction remains under-explored. A key distinction is that student modelling requires capturing individual learning trajectories and knowledge gaps over time, rather than reasoning about a single mathematical problem in isolation. Parameter-efficient fine-tuning methods such as LoRA~\cite{hu2022lora} have made it feasible to adapt LLMs to specific domains with limited computational resources. This has enabled a research direction of fine-tuning LLMs to solve the student prediction task, modelling the temporality of student answering in the LLMs context ~\cite{norris2026tokenknowledgetracingexploiting}. Despite this, systematic comparisons between LLMs and KT models across performance, latency, and deployment cost are lacking. Our work addresses this gap by benchmarking multiple LLMs, including GPT-4o-mini~\cite{openai2024gpt4omini}, Gemini-2.5-flash-lite~\cite{google2025gemini}, Qwen2.5-7B-Instruct~\cite{qwen2024qwen25}, Llama-1B~\cite{grattafiori2024llama3}, and a LoRA fine-tuned Llama-1B variant, against established KT models DKT~\cite{piech2015deep} and SAKT~\cite{pandey2019self} under identical evaluation conditions.

\section{Experimental Protocol}
\label{sec:experimental_protocol}
In this section, we describe our experimental setup: the task formulation and dataset, the models evaluated, the prompting strategies used for LLMs, and the infrastructure used for measuring latency and cost.

\subsection{Task Formulation}
The task we consider in this paper is a binary classification task where the models are tasked with predicting if a student will answer a future question correctly or incorrectly, given their answering history. The models have access to the question text or a question id and the construct text or construct id. This is shown in \Cref{tab:example_data}. In addition, for each question, we also have misconception text for each possible answer and explanations of the answers demonstrated in \Cref{tab:example_miscocneption_data}. To avoid cold-start issues for each model we took the decision to not make predictions on the first 10 questions. When predicting the answer to a question the model has access to the student's question-answer history. We decided on a representative load of 100k students requiring 40 answer predictions each to test the models on.
\vspace{-0.5cm}
\begin{table}[h]
\centering
\caption{Example Question and Construct Data}
\label{tab:example_data}
\begin{tabular}{lp{5cm}}
\toprule
\textbf{Component} & \textbf{Example} \\
\midrule
Question ID & 100 \\
Question Text & What is $2 + 3 \times 4$? \newline (A: 20, B: 14, C: 11, D: 26) \\
Construct ID & 10 \\
Construct Text & Order of Operations \\
\bottomrule
\end{tabular}
\end{table}
\vspace{-0.5cm}
\begin{table}[h]
\centering
\caption{Example Misconception and Explanation Data}
\label{tab:example_miscocneption_data}
\begin{tabular}{lp{5cm}}
\toprule
\textbf{Component} & \textbf{Example} \\
\midrule
Question Text & 6 pencils cost £ 1.50 How much do 3 pencils cost? \\
Answer Text & 0.75p \\
Misconception \\Text & Confuses decimal notation in money putting value in pounds but writing the unit for pence \\
Explanation \\Text & I think you have used the incorrect notation for money. Consider how the monetary values in the question are written \\
\bottomrule
\end{tabular}
\end{table}

\subsection{Datasets}
In \Cref{tab:dataset_stats} we detail some statistics about the datasets used throughout this work. The dataset is a real-world dataset extracted from an online learning platform. Notably, the training dataset is only used for the KT models or fine-tuned LLMs. The validation dataset is what is used to report model performance. For all datasets we restricted the number of responses for a student to 50, where the model is only tasked with predicting the final 40 question responses. The dataset bias is the percentage of correct responses. The students in the training and validation datasets are different and student data used to train the model is not used for validation. This means any trained models are being tested for their generalisability to new students. The questions used in the training and validation set are the same; there are less questions in the validation set due to the smaller dataset size and no student in that set having answered some questions.

\begin{table}[h]
\centering
\caption{Dataset Statistics}
\label{tab:dataset_stats}
\begin{tabular}{lrr}
\toprule
\textbf{Metric} & \textbf{Train} & \textbf{Val} \\
\midrule
\# Responses & 512,000 & 64,000 \\
\# Students & 12,800 & 1,600 \\
\# Questions & 4,252 & 4,104 \\
\# Responses per student & 40.0 & 40.0 \\
Dataset bias & 65.8\% & 66.5\% \\
\bottomrule
\end{tabular}
\end{table} 

\subsection{Metrics}
The two metrics we use to assess the predictive performance of the models are accuracy and F1 score. Accuracy is the most common metric used to measure model performance in KT model research. It provides an overall value for how often the model is correct and can be used to see if a model is beating the dataset bias. The equation for accuracy is given by
\begin{equation}
    \mathrm{Accuracy} = \frac{TP + TN}{TP + TN + FP + FN}.
    \label{eq:acc}
\end{equation}
The F1 score can be more valuable than accuracy when a dataset is not balanced, as it considers both how often a model successfully finds correct or incorrect answers via recall, and how often the prediction was correct via precision. In this work we consider a macro-averaged F1 score to assess how well does the perform across both correct and incorrect response prediction. The equation we use for F1 score is given by
\begin{equation}
    \mathrm{F1} = \frac{1}{2}\sum_{i=0}^{1}\frac{2TP_{i}}{2TP_{i} + FP_{i} + FN_{i}}.
    \label{eq:f1}
\end{equation}
In both equations $TP$ is the true positives where the model predicted the positive class correctly, $TN$ is the true negatives where the model predicted the negative class correctly, $FP$ is the false positives where the model incorrectly predicted the positive class, $FN$is the false negatives where the model incorrectly predicted the negative class, and $i$ is the class being considered. Here we deem class $0$ as the class of correct responses and class $1$ as the class of incorrect responses.

\subsection{Models}
Deep Knowledge Tracing (DKT)~\cite{piech2015deep} uses question ids and student answering history as model inputs and predicts if a student answers correctly or incorrectly on future questions. Self-Attentive Knowledge Tracing (SAKT)~\cite{pandey2019self} is a self-attention transformer-style model. SAKT similarly uses question ids and student answering history as model inputs and predicts if a student answers correctly or incorrectly on future questions. LLM KT is our custom-built encoder-decoder temporal transformer. But this model uses question, construct, explanation, and misconception text as model inputs and uses a Qwen 3 0.6B embedding model to map the text inputs into an embedding space that the KT model uses as an input. Importantly, we only use Qwen as a feature extractor to obtain fixed vector representations of questions, construct, explanation and misconception text. These embeddings are computed once and cached offline. At inference time, no LLM is involved in the actual student performance prediction. The temporality and modelling of student understanding is handled by our small, custom temporal transformer.

Gemini 2.5 flash lite~\cite{google2025gemini} and GPT 4o mini~\cite{openai2024gpt4omini} are both larger closed source LLMs. The specific details of these models are therefore not known but we can estimate their approximate size from online sources. For these models the student history is passed to the model through the input prompt. The input to the LLMs includes question ids, question text, construct ids and construct text. Llama 1B~\cite{grattafiori2024llama3} and Qwen 2.5 7B~\cite{qwen2024qwen25} are both open source LLM models. We similarly pass the student history to these models through the input prompt. For Llama 1B LoRA fine-tune we add a small percentage of extra parameters to fine-tune the model for the specific task of predicting students answers to future questions from their answering history. 

While the KT and LLM models operate on different inputs, they reflect the practical constraints and strengths of each model class.

\subsection{Prompting Strategies}
\begin{figure}[t]
     \centering
     \includegraphics[width=\columnwidth]{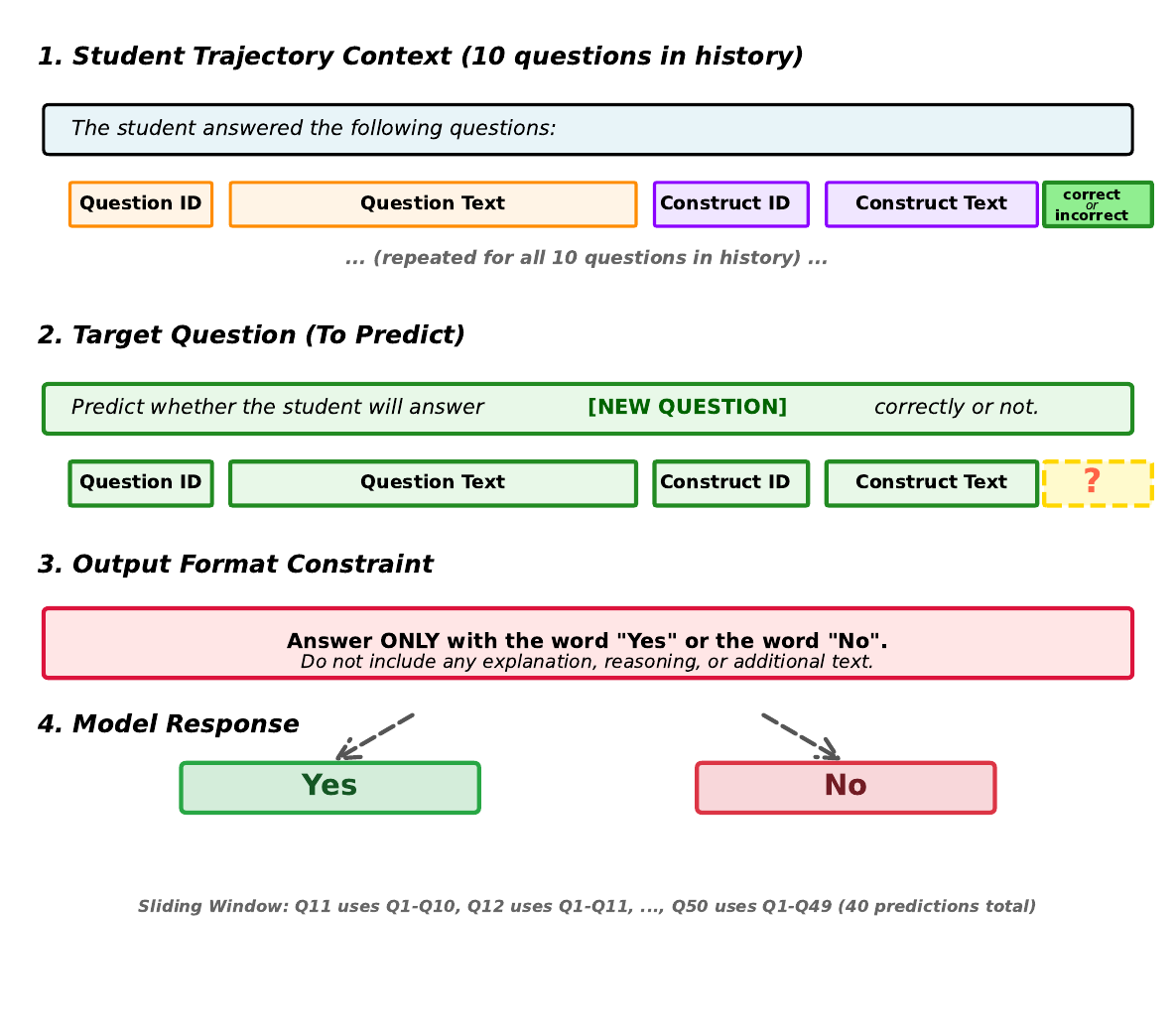}
     \caption{LLM prompt template for student response prediction.}
     \label{fig:prompt_LLM}
 \end{figure}

For student answer prediction, we employ a constrained prompt design that prioritizes deterministic parsing while providing rich contextual information.
Our prompt follows this structure: ``The student answered the following questions: Question ID [id] ([question text with choices A/B/C/D]), with construct ID [id] ([construct description]): answered [correctly/incorrectly]...
Predict whether the student will answer [new question with full context] correctly or not.
Answer ONLY with the word `Yes' or the word `No'.'' This is shown in \Cref{fig:prompt_LLM}.

\subsection{Testing Setup}
The testing of models is split into models that we would deploy ourselves and models that are closed source LLMs that we access via an API. For the models that we access via an API, we report the cost and latency values from using their API. For models that we deploy ourselves, we test the models on a CPU only Azure instance. This instance is a standard DS3 instance that has 4 vcpus and 14GiB of memory. We measure the latency of the model in a deployment setting where the model runs inference on one sample at once, although it should be noted that this could be sped up by increasing the batch size. We calculate the deployment costs based on an hourly rate of \$0.27 per hour of the instance.
\section{Results and Discussion}

\begin{figure}[t]
     \centering
     \includegraphics[width=\columnwidth]{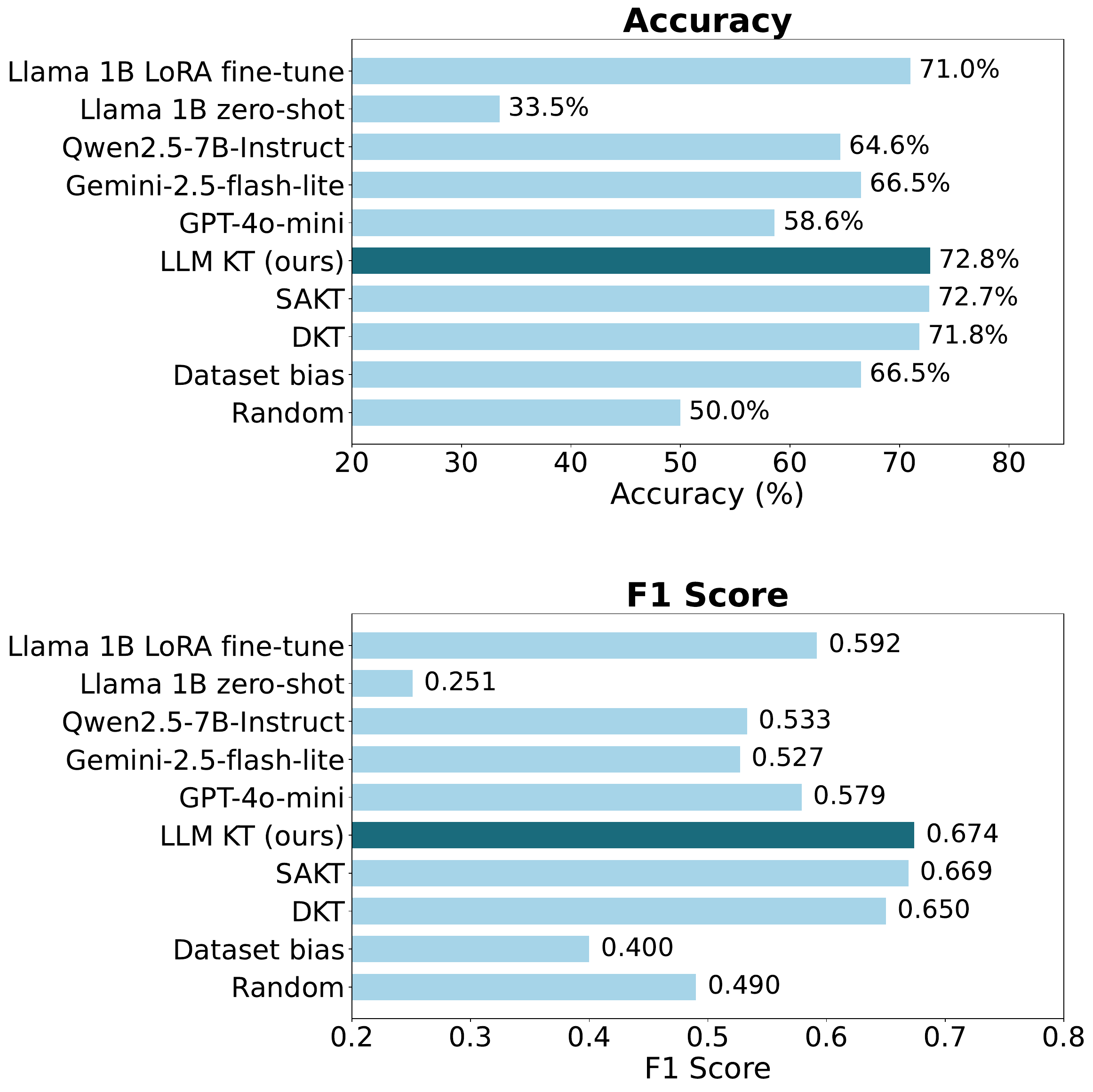}
     \caption{Comparison of model performance.}
     \label{fig:performance}
\end{figure}

In this section, we break down three key comparison points between KT models and LLMs for their use as predictive engines for EdTech platforms. We compare all models across predictive performance, latency, and deployment cost, which are key factors when deciding whether a model is suitable for a real-world use case. 

\subsection{Model Performance}


Accuracy measures the proportion of correct predictions. LLM KT achieves the highest accuracy at 72.8\%, narrowly outperforming both SAKT~\cite{pandey2019self} (72.7\%) and DKT~\cite{piech2015deep} (71.8\%), two well-established knowledge tracing models. Interestingly, all domain-specific KT models outperform the general-purpose LLMs like GPT-4o-mini (58.6\%), Qwen2.5-7B Instruct (64.6\%), and Gemini-2.5-flash-lite (66.5\%). We also benchmark two Llama-1B variants: a LoRA fine-tuned Llama-1B (71.0\%), while Llama-1B zero-shot performs substantially worse at 33.5\%. This reinforces the value of specialised models trained on structured educational data for this task. We refer the reader to \Cref{fig:performance} for a side-by-side comparison.

For context, we also include a dataset bias baseline that simply predicts the average correctness rate for each question (i.e., how often it is answered correctly in the dataset). While it does not personalise predictions, it achieves 66.5\% accuracy. Notably, several LLMs in our benchmark, despite their billions of parameters, fail to surpass this naive baseline. This makes it a critical baseline any viable model must beat.

Precision measures how often the model is correct when it predicts a student will answer correctly. Recall measures how many of the actually correct answers the model managed to identify. Improving one typically comes at the cost of the other. The F1 score captures this trade-off by combining both into a single measure, rewarding models that perform well on both rather than excelling at one while neglecting the other. The F1 score is therefore more informative when dealing with class imbalance. Once again, LLM KT leads with an F1 score of 0.674, followed closely by SAKT (0.669) and DKT (0.650). The LLMs perform lower in this metric as well, with GPT-4o-mini scoring 0.579, Qwen2.5-7B-Instruct 0.533, and Gemini-2.5-flash-lite 0.527. Among the Llama baselines, the Llama-1B LoRA fine-tune improves to 0.592, whereas Llama-1B zero-shot remains low at 0.251. These results affirm that general-purpose LLMs struggle with student-level predictions, likely due to the lack of fine-grained educational supervision.

 \begin{figure}[t]
     \centering
     \includegraphics[width=\columnwidth]{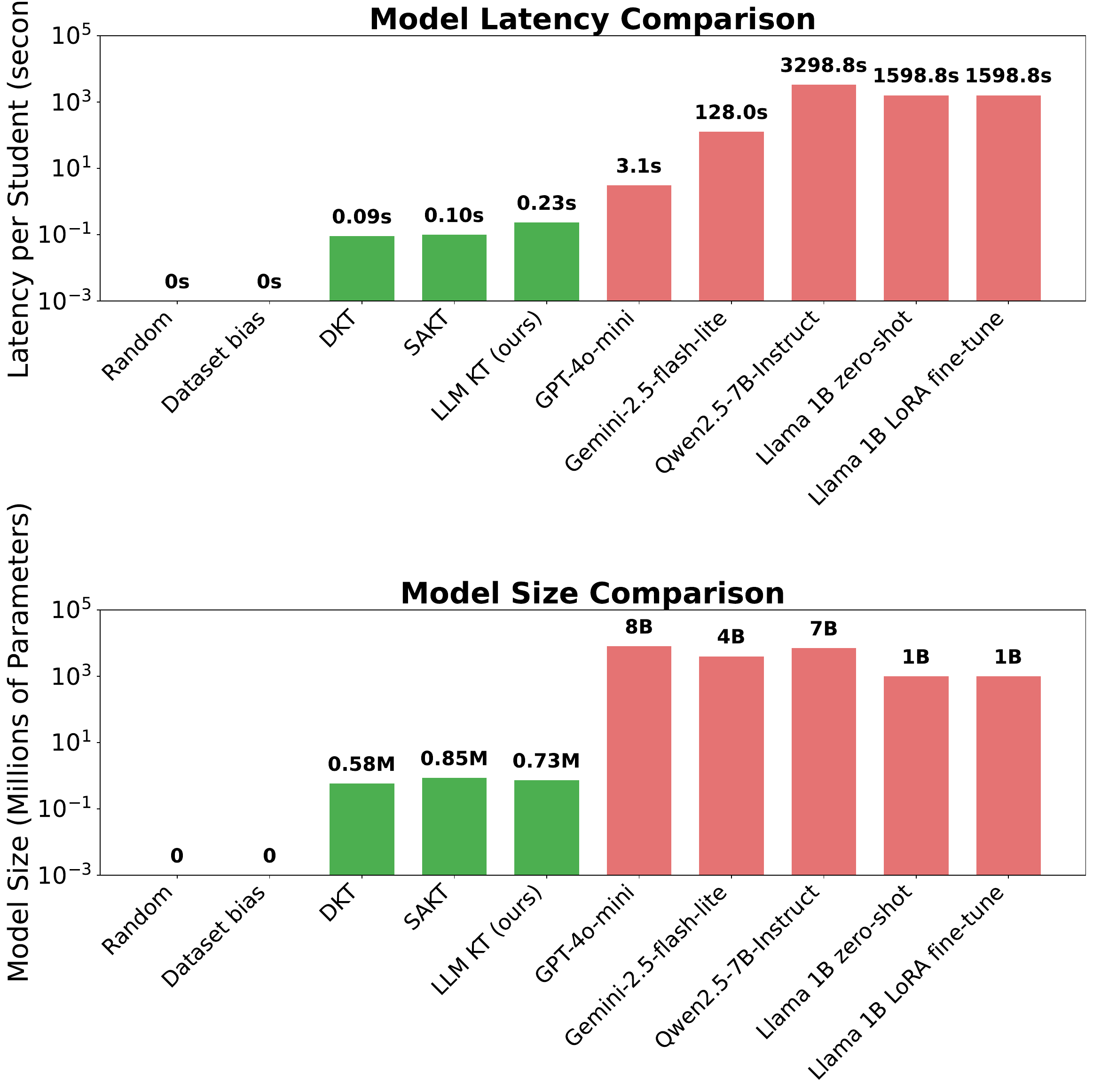}
     \caption{Latency and model size comparison for different models.}
     \label{fig:latency}
 \end{figure}

The LLMs exhibit different failure modes depending on their design. Llama-1B zero-shot frequently fails to follow the prompt format, producing responses other than the expected `correct'' or `incorrect,'' which we treat as incorrect predictions. The remaining LLMs (GPT-4o-mini, Gemini-2.5-flash-lite, and Qwen2.5-7B-Instruct, Llama-1B fine-tuned) adhere to the prompt format but struggle to predict student responses accurately. Notably, despite receiving ten initial question-answer pairs to mitigate cold-start effects, these models fail to infer students tend to get more answers correct in general. In contrast, specialised KT models are trained to leverage such properties of the data, enabling them to make more informed predictions about future student responses.

Our results show that specialised Knowledge Tracing models outperform general-purpose Large Language Models when it comes to predicting student responses. Across both metrics, domain-specific models like our LLM KT, SAKT, and DKT consistently deliver stronger performance than smaller LLMs such as GPT-4o-mini, Gemini-2.5-flash-lite, Qwen2.5-7B-Instruct, and Llama-1B. They remain robust even against a fine-tuned small LLM baseline (Llama-1B LoRA). This highlights the importance of task-specific architectures in educational prediction tasks. While LLMs excel at general reasoning, purpose-built KT models remain the most effective and reliable solution for identifying student misconceptions.

\subsection{Latency and Model Size}
We also compare models on latency (inference time per student) and model size (number of parameters) in \Cref{fig:latency}. These are critical factors for real-time applications and large-scale deployment in educational learning platforms. We note that closed-source LLMs run on different hardware and are accessed via an API. 

Traditional KT models like DKT, SAKT, and our LLM KT deliver fast predictions, with latencies under 0.25 seconds per student. In contrast, LLMs are orders of magnitude slower. GPT-4o-mini takes 3.1 seconds, Gemini-2.5-flash-lite takes 128 seconds, and Qwen2.5-7B-Instruct requires 3,299 seconds per student. Llama-1B variants also take 1,598.8 seconds per student. Despite this latency increase, none of the LLMs outperform the KT models in accuracy.

Specialised KT models are also extremely compact, with model sizes ranging from 0.58M to 0.85M parameters. Our LLM KT model contains only 0.73M parameters while offering strong performance. In contrast, general-purpose LLMs are vastly larger: GPT-4o-mini has 8B parameters, Gemini-2.5-flash-lite has 4B, Qwen2.5-7B-Instruct has 7B, and Llama-1B variants have 1B parameters.

This substantial efficiency gap in terms of latency and model size further reinforces that specialised KT models are far better suited for scalable, real-time prediction tasks. This is especially important for deployment in resource-constrained environments, where lightweight prediction models can enable educational tools to reach learners at a global scale.

\subsection{Cost Analysis}
\begin{figure}[t]
     \centering
     \includegraphics[width=\columnwidth]{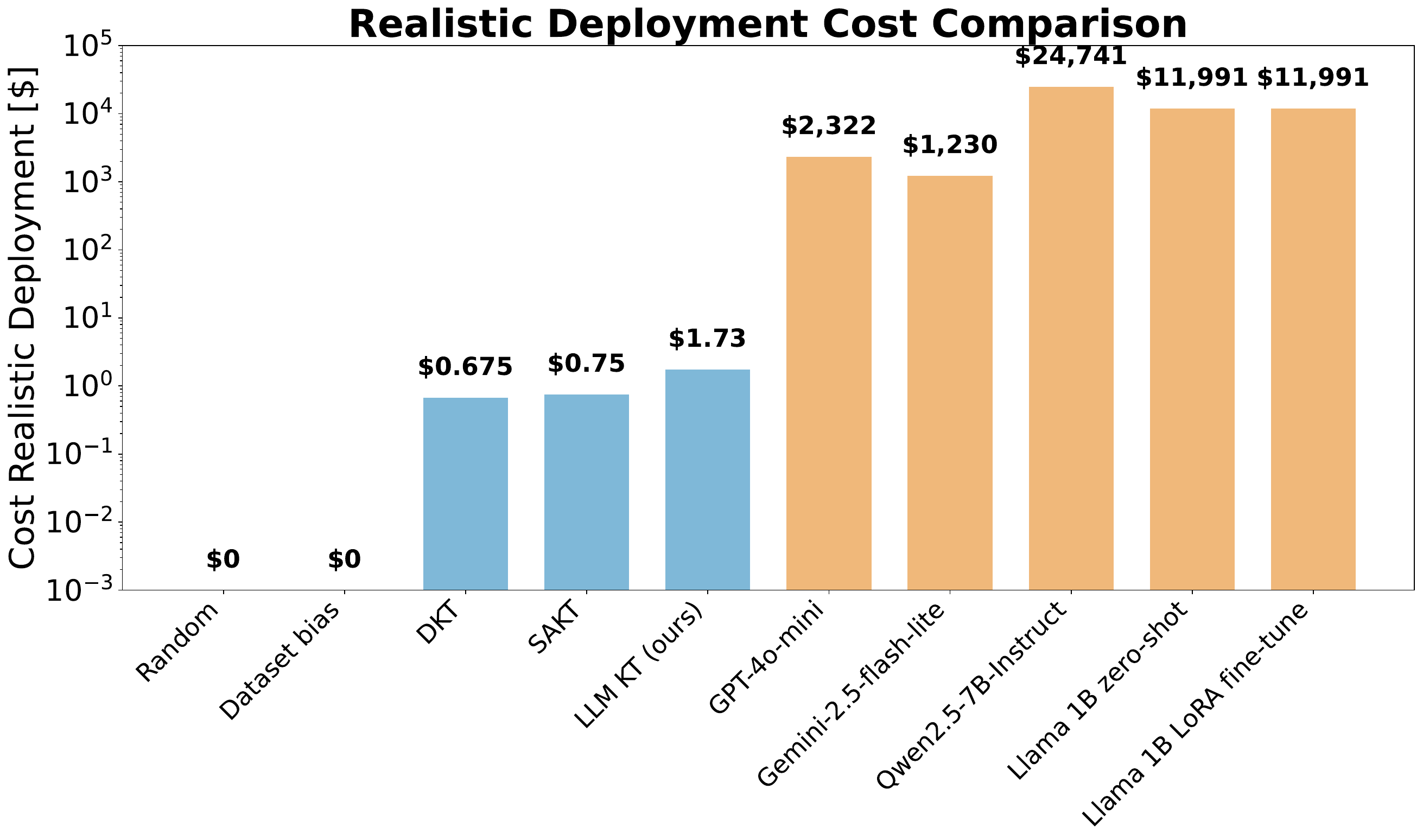}
     \caption{Annual inference cost comparison for 100,000 students receiving 40 predictions per year.}
     \label{fig:cost}
\end{figure}

Cost is a major consideration for real-world deployment at scale, especially for education platforms aiming to support thousands of students. Figure~\ref{fig:cost} shows the annual inference cost for 100,000 students, each receiving 40 predictions per year. For most LLM models, we benchmarked on 1,600 students with 40 questions each (64,000 predictions) and extrapolated to 100,000 students. For Gemini-2.5-flash-lite, daily rate limits of 10,000 requests per day on Tier-1 restricted us to 200 students, from which we extrapolated accordingly.

Specialised KT models like DKT, SAKT, and our LLM KT cost less than \$2 per year to serve this workload. In contrast, general-purpose LLMs are vastly more expensive: GPT-4o-mini costs approximately \$2,322 per year, Gemini-2.5-flash-lite costs \$1,230 per year, Llama-1B variants cost \$11,991 per year, and Qwen2.5-7B-Instruct reaches \$24,741 per year. This means KT models are 615--12,400 times cheaper than LLMs for the same task, while also offering higher accuracy.

For any large-scale EdTech deployment, this dramatic cost gap makes specialised KT models the better choice.

\begin{figure*}[htb]
     \centering
     \includegraphics[width=0.7\textwidth]{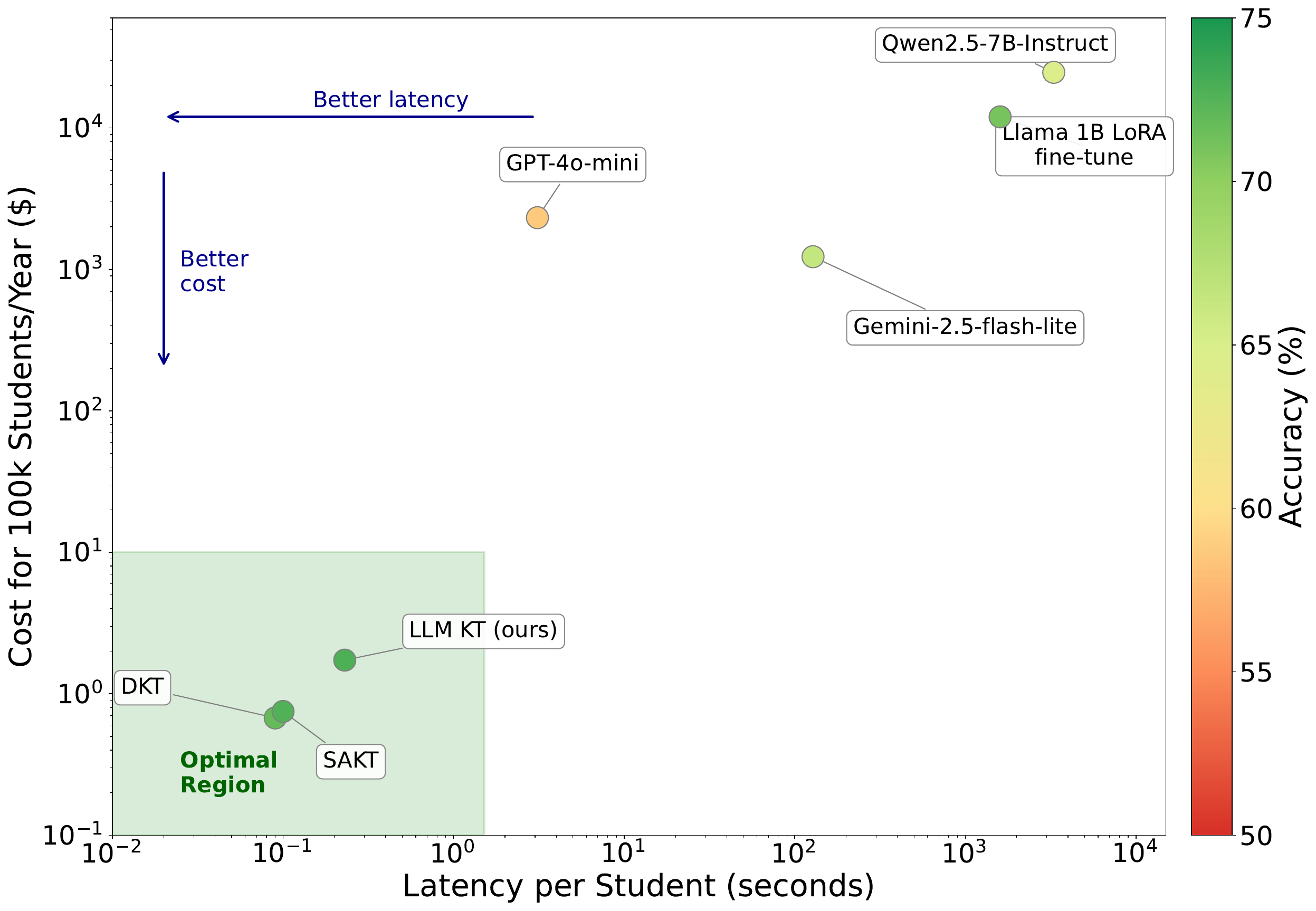}
     \caption{Latency vs.\ annual deployment cost for 100,000 students. Colour indicates model accuracy (green = higher, red = lower). Specialised KT models occupy the optimal bottom-left region.}
     \label{fig:tradeoff}
 \end{figure*}

 \subsection{Fast, Cheap, and Accurate: Why KT Models Outperform LLMs for Deployment}

To select a model for real-world use, we examine how latency, cost, and accuracy trade off at scale. Figure~\ref{fig:tradeoff} compares models on latency and deployment cost for 100,000 students per year. The x-axis shows latency per student (in seconds) and the y-axis shows annual cost for 100,000 students (in USD). Both axes use a logarithmic scale, so each step represents an order-of-magnitude change. Colour indicates accuracy using the scale on the right: greener points represent higher accuracy, while yellow, orange, and red points indicate lower accuracy. The optimal region is the bottom-left (fast and cheap) with green colouring (high accuracy).

Overall, the chart highlights an efficient frontier where specialised KT models achieve the best balance of speed, cost, and accuracy for real-world EdTech deployment.

\section{Conclusion}

Despite the rapid rise of general-purpose LLMs, our findings demonstrate that specialised KT models remain the best choice for predicting student responses in EdTech settings. KT models outperform LLMs (GPT-4o-mini, Gemini-2.5-flash-lite, Llama-1B, and Llama-1B-LoRA-finetune) on accuracy and F1 score, offering better future response predictions. KT models scale to larger numbers of students at a lower cost to EdTech providers due to being over 600 times cheaper to deploy at scale, while offering millisecond-level latency without requiring GPUs. While LLMs excel at broad reasoning tasks, they fall short when applied to student interaction data.

This is not to say LLMs have no place in pedagogical settings; however, they should be employed where they demonstrably reduce costs or improve learning outcomes, not as a default choice. To uncover misconceptions for the greatest number of learners, a scalable, real-time system is required. Currently, purpose-built KT models remain the most effective and practical solution.



%
\bibliographystyle{abbrv}
\bibliography{sigproc}  
%
\end{document}